\documentclass[letterpaper, 10 pt, conference]{ieeeconf}
\IEEEoverridecommandlockouts
\usepackage{cite}

\usepackage{graphicx}

\usepackage{amsmath}
\usepackage{amsthm}
\usepackage{amsfonts}
\usepackage{comment}

\usepackage[english]{babel}
\newtheorem{rec}{Recommendation}

\newtheorem{recconclusion}{Recommendation}

\usepackage{array}

\makeatletter
\newcommand{\thickhline}{%
    \noalign {\ifnum 0=`}\fi \hrule height 1.5pt
    \futurelet \reserved@a \@xhline
}
\newcolumntype{"}{@{\hskip\tabcolsep\vrule width 1.5pt\hskip\tabcolsep}}
\makeatother

\begin{document}
\bstctlcite{IEEEexample:BSTcontrol}

\title{\LARGE \bf Grounding Robot Navigation in Self-Defense Law}

\author{James Zhu$^1$, Anoushka Shrivastava$^2$, and Aaron M. Johnson$^1$ 
    \thanks{This material is based upon work partially supported by the U.S. National Science Foundation
    under grant \#CMMI-1943900.}%
    \thanks{$^1$Department of Mechanical Engineering, Carnegie Mellon University, Pittsburgh, PA, USA, \texttt{\{jameszhu,amj1\}@andrew.cmu.edu}}%
    \thanks{$^2$School of Computer Science, Carnegie Mellon University, Pittsburgh, PA, USA, \texttt{anoushk2@andrew.cmu.edu}}%
}

\maketitle
\thispagestyle{empty}
\pagestyle{empty}


\begin{abstract}

Robots operating in close proximity to humans rely heavily on human trust to successfully complete their tasks.
But what are the real outcomes when this trust is violated?
Self-defense law provides a framework for analyzing tangible failure scenarios that can inform the design of robots and their algorithms.
Studying self-defense is particularly important for ground robots since they operate within public environments, where they can pose a legitimate threat to the safety of nearby humans.
Moreover, even if ground robots can guarantee human safety, the perception of a physical threat is sufficient to justify human self-defense against robots.
In this paper, we synthesize works in law, engineering, and social science to present four actionable recommendations for how the robotics community can craft robots to mitigate the likelihood of self-defense situations arising.
We establish how current U.S.\ self-defense law can justify a human protecting themselves against a robot, discuss the current literature on human attitudes toward robots, and analyze methods that have been produced to allow robots to operate close to humans.
Finally, we present hypothetical scenarios that underscore how current robot navigation methods can fail to sufficiently consider self-defense concerns and the need for the recommendations to guide improvements in the field.

\end{abstract}

\section{Introduction}

Recent work in the robotics field has focused on deploying robots in public environments around humans for applications such as delivery \cite{lee_postal_delivery_robot_2021}, security \cite{lopez_robotman_2017}, and personal assistance \cite{vega-magro_robot_navigation_2020}.
Delivery robotics companies such as Starship, Kiwi, and Serve have been rapidly growing the number of robots operating in public spaces like university campuses and urban sidewalks.
In these environments, robots experience many close encounters with a diversity of humans.
It is unavoidable that a number of these encounters will result in some kind of danger to the robot and/or human, such as a robot combusting \cite{kiwibot_fire_2018}, college students vandalizing a robot \cite{haver_arrest_2022}, or a robot crashing into a car \cite{nbc_robot_crash_2020}.
Though uncommon, these edge cases underscore that humans and robots can, and will, come into direct physical conflict with each other.

While the examples above were a result of some combination of accident, negligence, and wanton violence, there are cases where a human using physical force against a robot could be justified under fear of threat to one's own well-being.
Consider the following scenario, which is discussed in further detail in Section \ref{sec:self_defense_justification}:
\begin{itemize}
    \item A person is tired after a long day of work and slowly walks home at night. A robot, mindful of keeping a safe distance from the person, quietly treads behind them at a constant distance. The person, on alert due to walking home alone at night, becomes fearful of something stalking them from behind. Eventually, the robot decides there is enough room on the sidewalk to accelerate and pass the person on their left.
\end{itemize}
In this example, both the human and robot behave reasonably, but the context and environment in which this interaction occurs could lead the human to feel threatened and act against the robot in self-defense.
Our research is focused on analyzing such situations where use of force in human-robot interaction may be justified under self-defense law.
In particular, we focus on a sub-group of people we anticipate to be the most likely to feel threatened by a robot: the non-expert non-user.
Compared to experts and/or users of robots, this group may not have an understanding of the capabilities that a robot possesses.

We propose categorizing self-defense scenarios as the tuple: (threat, protector, protectee), where the protector defends the protectee from some kind of perceived threat.
In standard self-defense where a human protects themselves from another human, this self-defense tuple is (human, human, self) where the protector and protectee are the same person.
There are also many other legally established tuples such as a human defending someone else from harm (human, human, other), a human defending their own property from someone else (human, human, property), and a human defending themselves from a non-human, which could be an animal or a non-living object (property, human, self).
Note that even though not all of these tuples represent a human defending themselves, we use the term self-defense in this work due to its broad familiarity.
This paper analyzes in detail the (robot, human, self) tuple, a special case of (property, human, self), while our future work will entail characterizing other self-defense situations involving robots.

We concentrate our analysis on public ground robots that have a primary task other than social interaction with humans.
These public robots contrast with industrial or care giving robots, where any person that interacts closely with them can be expected to have been trained or at least be quite familiar with how the robot operates.
Ground robots contrast with aerial robots like drones, which do not generally come into as close proximity to humans.

The objective of this paper is to first extract elements from prior works that are key to the formation of a (robot, human, self) self-defense scenario, and secondly present four actionable recommendations for roboticists to design systems that mitigate the likelihood and severity of these self-defense situations. 
These recommendations span research, industry, and policy making among the robots field, and were chosen based on their potential to progress the capability of robots to operate in dense public areas.
We aim to provide evidence-based guidance for roboticists across the industry so that the field can make positive, equitable impacts.

This paper draws upon the fields of self-defense law, human-robot interaction, and robot path planning to present an unaddressed topic that lies at their intersection.
These research areas are synthesized to establish how the current robotics state-of-the-art overlooks the possibility of non-expert, non-user humans acting in justified self-defense against a robot.
The rest of the paper is organized as follows:
\begin{itemize}
    \item Sec.~\ref{sec:self_defense_law} reviews the guiding principles that inform U.S. self-defense law and argues for the importance of considering (robot, human, self) scenarios.
    \item Sec.~\ref{sec:HRI} connects these principles to findings from human-robot interaction literature on human attitudes toward robots and identifies aspects of human attitudes that require additional study.
    \item Sec.~\ref{sec:human_aware_planning} discusses planning strategies that have been developed for robots to operate around humans and some of their limitations on handling self-defense situations.
    \item Sec.~\ref{sec:self_defense_justification} synthesizes the previous sections to present hypothetical scenarios where a human would be justified in taking self-defense action against a robot.
    \item Sec.~\ref{sec:conclusion} summarizes this work and discusses future work that the authors will address next to bolster the connection between robot design and self-defense law.
\end{itemize}

\section{U.S. Self-Defense Law}
\label{sec:self_defense_law}

In this section, we outline the key tenets of self-defense law in the United States.
These laws vary appreciably by jurisdiction in the extent and environments in which self-defense is justified.
For example, most U.S.\ states oblige no duty to retreat from a threat, while others impose a limited duty to retreat in public spaces \cite{brown1991no}.
Because of the great variance that exists in this domain, practitioners should use discretion with the specific self-defense statutes in their jurisdiction.
However, we believe the principles outlined in this section are broadly applicable to the vast majority of self-defense codes in both the U.S.\ and other countries, and the conclusions drawn in this work are largely independent of the specific intricacies of individual statutes.
The legal discourse regarding self-defense is lengthy and the analysis presented here is quite brief.
For a deeper discussion on the theory of self-defense law, see \cite{epstein_self_help_theory_2005,brandon_self_help_1984}.

The two primary principles that underpin American self-defense law are:
\begin{itemize}
    \item a reasonable belief of imminent physical harm
    \item a proportional response to the threat
\end{itemize}
While terse, these aspects of self-defense carry centuries of legal nuance that must be interpreted by the courts on a case-by-case basis.
Below we analyze some important details of these two principles.

\subsection{A Reasonable Belief of Imminent Physical Harm}
This phrase can be broken down into two parts. 
Firstly, a ``reasonable belief'' means that the protector in the (threat, protector, protectee) tuple does not need to have definitive proof that the threat of imminent harm is true, but only that it would be reasonable for a person to view a situation as threatening \cite{simons2008self}.
For instance, if the protector is acting with limited information about the threat, there can exist a reasonable belief of harm even if the protector is ultimately found to have been incorrect about the nature of the threat.
Similarly, the reasonableness of the belief is subject to prior experiences of the protector.
So even if an external third party would not necessarily view the belief as reasonable, context such as prior history of physical abuse \cite{scheff1997people} or legitimate verbal threats of violence that occurred directly prior \cite{state_v_sutton} can provide justification for acting in self-defense.

The second aspect of this definition is ``imminent physical harm'' \cite{rosen_imminence_1993}.
Imminence indicates that the harm to the protectee must be actively happening or about to happen.
This means that force used either proactively or after the harm has subsided is invalid to justify self-defense.
There has been debate centered around the soundness of imminence as an indicator of the necessity of self-defense, but in practice, this interpretation has been upheld \cite{rosen_imminence_1993}.
Additionally, the law tends to interpret the harm that is incurred as needing to be physical.
While some have argued that other types of harm such as invasion of privacy may be tantamount to a self-defense justification \cite{froomkin_self_defense_2015}, American common law has yet to grant this argument \cite{schneider2014open}.

\subsection{A Proportional Response to Threat}
Once the protector reasonably believes that self-defense is warranted, they must act in a manner appropriate to the threat level \cite{uniacke_proportional_2011}.
For self-defense between humans, this means that relatively minor force, such as a punch or kick, can not be responded to with lethal force.
However, certain situations can justify a lethal response to a less than lethal threat, such as when the protector is in their own home \cite{levin_castle_2010}.

In a (robot, human, self) scenario, it is key to understand that the robot is property and has no intrinsic right to act in self-preservation in the way that humans do.
Because of this, a self-defense act that destroys the robot can be justified even with a lesser threat to the human.
This concept of proportional response emphasizes the need for engineers to carefully design robots to avoid self-defense situations, because any perception of threat could lead to justified destruction of the robot.

\subsection{Self-Defense Against Robots}
Ground robots exhibit characteristics that can uniquely give rise to self-defense justifications.
While previous work has discussed self-defense against aerial drones \cite{froomkin_self_defense_2015}, the argument justifying self-defense against these technologies was weakened by the typically large distance between the drone and any given person.
Ground robots, on the other hand, are expected to come into close, immediate contact with humans during normal operation, so designing for self-defense situations is critical.

Since we have established that the proportionality criteria can be broadly satisfied in (robot, human, self), the primary challenge to determining when people may be justified to defend themselves against a robot is defining what a reasonable belief is.
Because robots are still such a novel and unfamiliar technology to most people, the standard of reasonableness may be lowered to take into account the misconceptions and misunderstandings non-experts tend to carry as they interact with and react to robots.

To begin codifying self-defense law as a serious consideration for robot engineers and to assuage the public's fears about coming into contact with potentially dangerous robots, we make the following recommendation:
\begin{rec}
Robotics companies and research organizations should publicly advertise that in situations where a human and robot are in direct physical conflict, the human's physical well-being is always valued more greatly than that of the robot, even if the result is damage or destruction of the robot.
\label{rec:proportionality}
\end{rec}

This recommendation reinforces the justification of the proportionality criteria for self-defense against robots, which will empower people with the understanding that they will have at least some immediate recourse if they feel physically under threat by a robot.
In turn, this transparency from the robotics field on the rights that people have when interacting with robots can engender a greater sense of trust and openness to the deployment of robots in public spaces.
As discussed in Section \ref{sec:HRI}, the attitudes that humans have towards robots has a large impact on how a human-robot interactions play out.

\section{Reasonable Perspectives Toward Robots}
\label{sec:HRI}
In this section, we take a step toward understanding standards of reasonableness in human-robot interaction. 
We start by analyzing how behavioral norms can dictate (human, human, self) self-defense scenarios. 
We next contrast the role of behavioral norms in human-human and human-robot interactions and present results from prior literature on human attitudes toward robots, which contribute to an understanding of reasonable behavior. 
There lacks consistently interpretable robot behaviors that could be used to establish human-robot norms, and instead suggest considering diverse attitudes toward robots to inform what constitutes reasonable behavior.

\subsection{Behavioral Norms in Human Interactions}

In human-human interactions, behavioral norms between humans play a crucial role in determining reasonable behavior.
When evaluating (human, human, self) cases, courts judge the reasonableness of a person's behavior based on broad, implicit understandings of how humans typically behave. 
For example, in the case of Rowe v. United States, it was found that even though Rowe kicked a man, which prompted the man to attack him with a knife, Rowe stepping back after the kick revived his right to self-defense \cite{kopel_self_defense_2000}.

Following the literature \cite{anderson2014behavioral}, we define a norm as a widely adhered to and understood action that helps coordinate behavior. 
This stepping back is a norm that was found to be a widely understood indication that Rowe was no longer a threat and withdrawing from the fight, which protected him from being retaliated against. 
Such implicit understandings between humans can be considered a basis of self-defense law because they create standards of reasonableness, allow for generalizations across cases, and ultimately promote fairness in decision-making.

Norms for human-robot interaction, on the other hand, are not well established in the sense that actions taken by the robot are not widely understood and do not facilitate coordinated behavior between the human and robot, which can make robot behavior unpredictable to a non-expert. 
Even though human behavior is not completely predictable, years of lived experience interacting with other humans allows for a deeply developed understanding of typical human behavior. 
Due to the novelty of robots, this understanding between humans and robots is inadequate.
Because of the lack of established norms for robots, the standard for justified self-defense in (robot, human, self) should be lower than in (human, human, self) cases.

Frameworks such as COMPANION have attempted to address this gap by encoding human social norms (such as maintaining personal space and moving to the right to avoid colliding with people approaching from the opposite direction) within robot behavior \cite{kirby_companion_2009}. 
However, more research is needed to determine whether humans actually expect robots to behave with the same norms as humans. 
In fact, some studies have indicated otherwise. 
For example, \cite{sardar_stand_2012} found that robots were considered more trustworthy when approaching a person quickly (possibly since faster robots were more noticeable than slower robots), whereas humans were more trustworthy when approaching slowly. 
The study also found that humans performed more corrective reactions (such as stepping back or adjusting eye contact) when a robot invaded their personal space compared to a human.
This indicates that humans react differently to violations depending on if the offending party is a human or robot, and simply having robots adopt human norms does not guarantee self-defense situations will be avoided.

Even if robot behavioral norms become standardized among the industry, there is no guarantee that humans will, in their split-second decision making, have enough trust to assume that a robot can reliably follow certain norms.
It is important for robots to conform to the preferences and expectations that humans have for their behavior, and we recommend further study of these topics:
\begin{rec}
    Because of the differences in human preferences and expectations when interacting with robots compared to other humans, researchers should explore whether there are robot behaviors that humans react consistently to and if these behaviors can be encoded into a standardized framework of human-robot norms.
    \label{rec:norms}
\end{rec}

Results that establish norms for even a subset of ground robots (such as wheeled robots, humanoids, or quadrupeds) could begin to establish a more refined definition of reasonable behavior around robots.
Possibilities of behavioral norms for robots could be exhibiting body language or digital facial expressions \cite{cohen_childs_recognition_2011}, which could in turn improve the legibility of robots in public environments (as discussed in Sec.~\ref{sec:legibility}).
Establishing norms may be difficult due to the variance in attitudes that humans exhibit toward robots.
Even if consistent norms cannot be identified for many aspects of robot behavior, it is still important to characterize how attitudes vary among humans and under what circumstances human attitudes can be well modeled.

\subsection{Human Attitudes Toward Robots}
Human attitudes towards robots tend to vary based on several factors, including a person's familiarity with robots and how well a robot's behavior aligns with the human's expectations and preferences. 
Studies suggest that the more familiar a person is with a robot and the more their expectations align with the robot’s behaviors, the more positive their attitude towards that robot \cite{strait_public_perception_2017}.
Conversely, when there are gaps and discrepancies in these areas, attitudes tend to shift negatively.
The real consequences of negative attitudes toward ground robots and a violation of expectations during their deployment emerge as justified self-defense scenarios.
Alleviating the public's negative attitudes and aligning robot design with expectations is essential for safe human-robot interactions.
This reinforces the necessity of Recommendation 1, which can help inform the public's expectations of how they can interact with robots.

One highly-documented example of this is the largely negative attitudes communities have expressed toward the deployment of robots by police departments in several U.S.\ cities \cite{yunus_responsible_use_2021}.
One issue that arose was that communities affected by the police's usage of ground robots were not involved in the development process and expressed frustration over the expensive and possibly dangerous technology \cite{yunus_responsible_use_2021,zaveri_nypd_robot_dog_2021}, calling the robot ``another danger for Black \& Latino residents'' \cite{zaveri_nypd_robot_dog_2021} and expressing fear toward the futuristic appearance of the robots \cite{periwal_robot_strikes_fear_2020}.
Requests to police departments for more information on the purpose of the robots were not always met \cite{yunus_responsible_use_2021,jarmanning_spot_advocates_2019}.

When impacted communities are not involved in the development of robots, the deployed products can be misaligned with the community's expectations of how these new tools should be used. 
Research indicates that the level of familiarity people have with robots and the preconceptions they hold influence their attitudes, such as how the fear of sentient robots correlates with negative attitudes \cite{strait_public_perception_2017}. 
Another study suggests that people may be more likely to support robots doing jobs that require less experience and communication \cite{takayama_what_robots_should_do_2008}. 
Therefore, attitudes toward delivery robots, which satisfy both of these conditions, could be more positive than those toward police robots, though further research should test this theory.


Attitudes of marginalized groups toward robots are especially important for developers to consider, since they have been disproportionately affected by harmful uses of novel technologies \cite{buolamwini_2018}.
Police robots have often been deployed to patrol low-income, Black neighborhoods \cite{yunus_responsible_use_2021,gidaris2021rise}, while women have repeatedly been targets of unwelcome surveillance by drones \cite{thomasen_feminist_2018,tourjee_stalked_2017}.
Additionally, \cite{gurinskaya_robocops_2022} found that women tended to be less receptive to the concept of patrolling police robots than men. 
To combat this inequity, \cite{yunus_responsible_use_2021} suggests involving marginalized community members in the technology design process. 
Factoring in the preferences of the stakeholders who interact most closely with robots will help developers align robot design with expectations, reduce negative attitudes toward robots, and promote equity by working for marginalized communities instead of against them. 
We recommend further investigation into how attitudinal differences manifest among disadvantaged groups:

\begin{rec}
    Due to the variance in human attitudes toward robots and the disparate effects technologies have had, researchers should examine and catalog the attitudinal differences among different groups of people, especially from those that have historically been marginalized.
    \label{rec:attitudes}
\end{rec}

Reasonable human behavior varies greatly due to differences in background and past experiences. 
These differences can be measured in the attitudes, perspectives, and reactions people exhibit toward robots. 
Ultimately, this variability suggests that even severe human behaviors towards robots can be justified and considered reasonable, at least until the establishment of robot behavioral norms that are broadly understood by people of many backgrounds. 
Instead of basing robot behavior on unestablished norms that people must adhere to, it is essential to consider people's diverse attitudes and expectations regarding robots and design robots with this context in mind. 

\section{Human-Aware Planning}
\label{sec:human_aware_planning}

While self-defense has so far gone unconsidered in the design and implementation of robots, there has been ample related work in planning robot motions in human environments.
Generally, robot path planning is performed by sampling many possible paths a robot can take and selecting the most optimal choice, often based on the shortest path \cite{hart_a_star_1968,lavelle_rrt_1998}.
Algorithms are also able to obey specified constraints such as avoiding obstacles.
Recent research has adapted these path planning algorithms to predict and react to human obstacles, and to minimize risk of collision with people \cite{chu_risk_aware_2021}.
Other work has developed robotic behaviors to satisfy desired outcomes such as visibility \cite{taylor_legibility_2022}, active communication \cite{sanoubari_nonverbal_2022}, and following social norms \cite{chen_motion_planning_2017,kirby_companion_2009}.
In this section, we examine two primary research thrusts in human-aware planning that have seen significant attention: explainability and legibility.
We analyze not only what work has been done, but also the reasons stated in the literature for why these aspects of human-aware planning are important.
While aspects of explainability and legibility are useful in mitigating the potential for self-defense situations, current implementations lack the capability to address all environments in which a self-defense scenario may arise.

\subsection{Explainability}

Drawing from \cite{langley_explainable_agency_2017}, we define explainability as the ability of an autonomous agent to produce records of the decisions it has made and understandable reasoning for why those decisions were made.
This definition is compatible with how explainability is discussed in prior works, such as generating contrastive explanations (i.e.\ why A and not B?) \cite{cashmore_explainable_ai_2019} and explanations that satisfy user-defined preferences \cite{ sohrabi2011preferred}.
Post-hoc explanations are designed to be generated after a robot decision has been executed and in response to some kind of questioning, while some work has examined generating concurrent explanations for behaviors as they are happening~\cite{zhu_proactive_explanations_2020}.

Drawing from the existing discourse on explainability in the field of autonomous vehicles (AVs), \cite{atakishiyev_explainable_av_2021} discusses post-hoc and concurrent explanations for AVs by analyzing a scenario in which an AV fails to recognize a pedestrian crossing in front of it.
Post-hoc explanations to characterize why the AV failed could be useful in a post-accident investigation and for regulators to hold manufacturers accountable.
But these post-hoc explanations would not be able to prevent accidents from occurring.
Concurrent explanations, such as communication to a passenger that the car will continue through a crosswalk because no pedestrian has been detected, could allow passengers to take emergency actions when they recognize the vehicle has made an error.
In this instance, a passenger could activate an emergency brake that stops the car before it collides with the pedestrian.

In the context of ground robots, post-hoc explanations assume that the people who desire explanations for robot behaviors have access to the robot afterward.
These explanations could be useful to operators who could recognize errors in their usage of the robot, developers who could better understand errors and implement fixes, and members of the judiciary who could use explanations to assign liability after an accident.
However, post-hoc explanations generally exclude members of the general public who interact with the robot for just a fleeting moment, such as passing each other on the sidewalk.
Considering that the majority of people interacting with a robot in a public environment will likely not have access to that robot afterward, this exclusion is significant.
Concurrent explanations, on the other hand, are able to actively communicate to people in the robot's immediate surroundings.
However, concurrent explanations may be difficult to convey to certain people in real-world environments.
Explanations announced verbally may not be heard by people on the phone or listening to music, or may be drowned out in loud environments such as construction.
Similarly, explanations presented visually may not be suitable for people with visual impairments or in night-time environments.
A robot must also consider that some people may not speak the robot's default language.

In a survey of 62 papers on explainability, \cite{anjomshoae2019explainable} found that the most commonly stated motivation for the work was transparency (i.e\ allowing people to better understand the inner workings of the robot), followed by trust and collaboration.
These motivations go hand in hand, as increased transparency would naturally lead people to trust being around the robot and working with it.
Of these surveyed papers, many framed trust around the relationship between robots and their operators or teammates, and the faith these people had that their robots would work reliably \cite{boyce_transparency_2015, wang_trust_2016}.
With the European Union's recent General Data Protection Regulation (GDPR) outlining a person's legal ``right to explanation'' when encountering autonomous agents \cite{eu_gdpr_2016}, explainability has come to even greater relevance.

In this work, we question if explanations are the key to inspiring public trust in robots and allowing for a seamless deployment of robots in public spaces.
Explanations can be valuable, but are ineffective in fully mitigating the possibility of self-defense scenarios due to the difficulties of access and communication with typical bystanders.
What robots need is the capability to generate implicit methods of communication that can foster an improved understanding of robot behavior.
The human-aware planning concept of legibility may be a more suitable method to accomplish this.

\subsection{Legibility}
\label{sec:legibility}

While explainability focuses on the producing reasons for why a robot behaved a certain way, legibility characterizes how a robot communicates what it is doing or intends to do. Based on the work from \cite{sisbot_motion_planner_2007}, we define legibility as the ability of a human to understand a robot's intentions based on observation.
For instance, cars have turn signals to indicate to others what action they are about to perform (e.g.\ turning right).
The turn signal does not explain why the car is turning right, but allows others to understand what it is about to do and react accordingly.
Explanations can help robots become more legible, but there are many other factors that can improve legibility such as providing cues \cite{hetherington_cueing_2021}, mimicking human behavior \cite{guzzi_human_friendly_2013}, moving quickly toward goals \cite{dragan_legibility}, and staying in people's field of view \cite{taylor_legibility_2022}.

Legibility is a somewhat vague concept that is difficult to define and measure experimentally.
To do this, some authors have evaluated legibility by asking people to predict a robot's future behavior based on past observed behavior \cite{busch_learning_legible_2017,wallkotter2022new}.
Others evaluated human performance of an unrelated task while a robot navigated around them \cite{bortot_human_approximation_2013}.
Others still used questionnaires to gauge how well subjects felt they understood a robot's intentions \cite{hetherington_cueing_2021}, while \cite{dragan_legibility} proposed a numerical measure of a trajectory's legibility.

Studies have found that legibility is correlated with increased feelings of safety, comfort, and acceptance \cite{bortot_human_approximation_2013, eyssel_anticipation_2011,dehais_human_safety_2011}.
These goals align very closely with the stated purposes of explainability, but may differ in the groups of people these methods are designed for.
While explainability is often framed around expert users, significant work has focused on robot legibility to non-experts, as the lesser amount of information needed may make it easier for non-experts to comprehend.
However, as with generating explanations, interacting with a diverse group of humans may make producing legible behaviors much more difficult.
In these situations, it may not be possible to stay in everybody's field of view or to expect all people to notice the robot's gestures.
The complicating factors discussed for explainability can make communication difficult here as well, as visual or verbal signals can breakdown in certain cases.
Engineers must also consider how obvious any given message is to non-expert non-users.
This relates closely to the concept of human-robot norms, discussed in Sec. \ref{sec:HRI}.
Norms are not yet established for robots, so developing robots that can be legible to people across diverse backgrounds is an enormous open problem. 

\subsection{Human-Aware Planning for Non-Expert, Non-Users}
Based on this section, we conclude that current human-aware planning algorithms have not sufficiently addressed how robots should operate in dense human environments where communication is impeded.
In particular, non-expert, non-users have received little attention as to how robots should interact with them.
For instance, the IEEE Standard for Transparency of Autonomous Systems lays out guidelines for what information robots should be able to communicate to people \cite{ieee_transparency_standard_2022}.
The transparency standard for users is grounded primarily in providing explanations, but the transparency standard for non-users among the general public is focused exclusively on data privacy and lacks considerations for how robots must communicate with non-users.
Other standards associations like ISO and ANSI address robots in industrial, service, or personal care environments, but also do not acknowledge non-expert non-users in public settings.
We recommend these standards be revised to acknowledge how non-expert, non-users interact with robots:
\begin{rec}
    Standards, guidelines, and regulations from influential organizations such as IEEE should detail how robots should interact with non-expert, non-users to cultivate perceived safety and trust amongst the public.
    By drawing from concepts such as legibility, these standards can provide clear direction for how robots should be developed to minimize self-defense occurrences.
    \label{rec:standards}
\end{rec}

While current standards for transparent and legible robots are unsatisfactory, the literature provides guidance to suggest possible standardization of certain aspects of robot behavior, such as establishing a standard mapping of light color to indicate behavior \cite{froomkin_self_defense_2015,thomasen_feminist_2018} and requiring the deployment of noticeable robots instead of silent, stealthy ones \cite{sardar_stand_2012}.
Another regulation that could be considered is how robots should operate at night.
The Federal Aviation Administration currently restricts flying drones at night \cite{froomkin_self_defense_2015,FAA_night_drones} and encouraging similar regulations for ground robots could be productive. 
However, unlike drones, there is not one government agency that can dictate regulations on ground robots.
State and local governments control the use of their roads and sidewalks, so adoption of a consistent set of regulations is unlikely.
Organizations like IEEE and leading companies like Boston Dynamics and Agility Robotics may be the vanguards of establishing industry standards for operating robots.
Regardless, there is still a long way to go in establishing standards and regulations that fully address self-defense against robots.

\section{Justified Human Self-Defense Against Robots}
\label{sec:self_defense_justification}

Even when explainability and legibility are incorporated into robot planning, there is still potential for self-defense situations to arise between robots and non-expert, non-user humans.
Practitioners must understand the dependence a human-robot interaction has on the exact person and environment the interaction takes place in.
In the Introduction, we suggested a hypothetical case of a lone person walking home at night, who is followed from behind and then passed by a robot.
Consider two additional hypothetical cases where a robot that behaves according to conventional human-aware planning principles makes a human feel uncomfortable and even threatened:
\begin{itemize}
    \item A robot equipped with gaze tracking technology attempts to stay as close as possible to the center of a person's field of view to maximize legibility as they wait for a bus. 
    This person, however, is attempting to view the numbers of the buses that are passing by, and is unsettled that the robot appears to be blocking them from finding the right bus to leave on.
    As the robot approaches, they become afraid of the robot's single-minded focus on them.
    \item Navigating around a blind corner, a legged robot unexpectedly bumps into a person turning the corner in the opposing direction.
    In this situation, the robot is unable to satisfy the personal space constraint it has been programmed with and reverts to a ``safe mode'', which is to sit down on the ground.
    Already flustered by the sudden encounter with the robot, the person finds this behavior particularly unexpected and feels unsafe due to this unpredictability.
    
\end{itemize}

In each of these cases, the robots demonstrate some aspects of current human-aware planning methods, which in many circumstances may be appropriate and increase the transparency, trust, and perceived safety that nearby people feel.
However, these cases highlight ways that naive implementation of these methods can cause unintended negative effects.
The robot in the first case from the Introduction takes care to maintain a safe distance and pass according to typical social norms, but fails to account for the context and environment that causes the person to be fearful of any nearby entity approaching quietly from behind.
In the second case, the robot attempts to maximize its visibility, but without the understanding that the human would prefer to not have the robot so central in their field of vision and feels uncomfortable with the intense attention the robot is paying them.
Finally, the third case highlights a robot's attempt to embody a norm that indicates a non-threatening disposition.
However, this norm is not obvious enough to a person that must make a split-second decision on whether the robot could harm them.

Even though the robots in these examples may not pose an actual threat to the humans, the behaviors of these robots coupled with the people's backgrounds and the unique environment they are in can lead to a perception of threat.
This perceived threat could manifest into the humans acting in self-defense against the robots once they are sufficiently close to each other.
Self-defense in these (robot-human-self) cases would be justified because there exists a reasonable belief of imminent physical harm.
These scenarios could result in damage or destruction of the robot, a potentially appropriate proportional response to the threat.

\section{Discussion and Conclusion}
\label{sec:conclusion}

As roboticists work to rapidly ramp up deployment of their robots in public environments, it is crucial to understand the genuine physical harm these robots could cause human bystanders.
Developers must design robots not only to guarantee human safety, but also to maximize the perceived safety of nearby humans.
However, as robots are still largely unfamiliar to most of the general population and are often viewed with negative preconceptions, it is likely that some humans will see robots as threats to their physical safety and act in self-defense.
In this work, we discuss how self-defense law applies to human encounters with ground robots, the human norms and attitudes that dictate the outcome of human-robot interactions, and the need to expand explainability and legibility to address self-defense cases.
Synthesizing these three concepts, we identify scenarios where human self-defense against robots could be justified, even under reasonable robot behavior.

These considerations inform four recommendations to roboticists that aim to reduce the likelihood of justified human self-defense against robots.
Recommendation 1 addresses robotics companies and research institutions to provide open communication to the public on the rights that they have when interacting with robots they perceive as dangerous, which will promote public trust and improve human attitudes toward robots.
Recommendation 2 suggests researchers examine if any implicit robot behaviors are widely interpretable to humans and if a framework of human-robot norms can begin to be established.
Recommendation 3 calls for a more detailed exploration of the attitudes that marginalized groups such as Black communities and women hold toward robot deployment.
This will work toward considering previously-excluded people in the development of novel technologies and reinforcing the rights of these marginalized populations.
Finally, Recommendation 4 advocates for an overhaul in robot standards, guidelines, and regulations to address legible robot behavior to non-expert, non-users.

We argue that contextualizing robot navigation in self-defense law establishes tangible, relevant outcomes that developers can use to evaluate their algorithms on.
We hope that this work will contribute to keeping people of all backgrounds safe and secure as robots are increasingly deployed around them.

Based on Recommendations \ref{rec:norms} and \ref{rec:attitudes}, one line of future work we would like to explore in more detail is the reasonableness criteria for self-defense against robots.
Specifically, we would like to experimentally establish the aspects of robot locomotion that people would be more or less likely to perceive as threatening.
The gaits of legged robots in particular are an unexplored aspect of human-robot interaction we aim to investigate.
We would also like to expand the (threat, protector, protectee) framework by exploring scenarios where robots could act as protectors.
The legal justification for such a scenario may be much more narrow than humans protecting themselves, but some work has begun to look at robots defending themselves \cite{zhu_drones_2022} or their human users \cite{duarte_robot_foce_2022}.

\addtolength{\textheight}{-5.5cm}   
                                  
\bibliographystyle{IEEEtran}
\bibliography{ref}

\end{document}